# Data-driven project planning: An integrated network learning, process mining, and constraint relaxation approach in favor of scheduling recurring projects

Izack Cohen

***Abstract*—Our focus is on projects, i.e., business processes, which are emerging as the economic drivers of our times. Differently from day-to-day operational processes that do not require detailed planning, a project requires planning and resource-constrained scheduling for coordinating resources across sub- or related projects and organizations. A planner in charge of project planning has to select a set of activities to perform, determine their precedence constraints, and schedule them according to temporal project constraints. We suggest a data-driven project planning approach for classes of projects such as infrastructure building and information systems development projects. In such projects, a significant portion of activities recurs within other organizational projects, which may be similar, while each project is unique in its realization. The first steps of the suggested approach include learning a project network from historical records of similar projects. The discovered network relaxes temporal constraints embedded in individual projects, thus uncovering where planning and scheduling flexibility can be exploited for greater benefit. Then, the network, which contains multiple project plan variations, is enriched by identifying decision rules and frequent paths in favor of selecting a specific variation as the chosen project plan. The planner can rely on the suggested approach for: 1) Decoding a project variation such that it forms a new project plan and 2) applying resource-constrained project scheduling procedures to determine the project's schedule and resource allocation. Using two real-world project datasets, we show that the suggested approach may provide the planner with significant flexibility (up to a 26% reduction of the critical path of a real project) to adjust the project plan and schedule. We believe that the proposed approach can play an important part in supporting decision making toward automated data-driven project planning.**

## I. Introduction

Projects are replacing operations as the economic driver of our times. In Germany, for example, projects accounted for 41% of the GDP in 2019. It is estimated that global project-oriented economic activity will reach $20 trillion in 2027 with 88 million people working in project management-oriented roles [1].

Differently from operational processes such as services (e.g., banking, retail, medical services, call centers etc.), which are performed by pools of organizational resources without detailed planning [2], projects are constrained by contractual obligations and demand significant time and cost investments. They also have higher complexity and uncertainty levels than operations [3] and thus require detailed planning, resource allocation, scheduling, and control. Binding due dates and milestones are typically associated with penalty/award mechanisms that underscore the importance of detailed, high quality project planning and execution.

This paper proposes a data-driven project planning approach that learns from past projects, revealing activity patterns and decision rules, and relaxing redundant constraints, all of which can enhance the project modeler's (i.e., a planner's) capabilities.

We focus on so-called non-unique projects such as construction projects, aircraft refurbishment and maintenance projects, and information systems development projects. In such projects, a significant portion of activities recurs within other similar organizational projects, yet each project is unique in its realization. In other words, projects of the same type (e.g., a 737-400 aircraft C-check) are likely to have many similar activities although some activities, activity sequences, and their durations may be different. Adler et al. [4] who studied such projects stated that "...while projects are often managed as unique configurations of tasks, in reality different projects within a given organization often exhibit substantial similarity in the flow of their constituent activities". For more information about characterizing non-unique projects, see [5].

The PMBOK Guide [6], the most popular project management standard today, teaches that the preliminary steps before scheduling a project are to define its activities and then to sequence them, after which a project network that presents the relationships between activities can be prepared. To this end, the PMBOK Guide offers techniques such as acquiring expert judgement, holding meetings, precedence diagramming, and establishing a project management information system. These techniques depend heavily on experience and time-consuming manual labor. Indeed, due to the scale and complexity of projects, which include dozens to hundreds of linked activities, a planner would typically opt to create a new project plan based on a plan from a similar previous project and modify it to meet the new project's requirements.

In fact, analogy-based planning was the common practice in a large aerospace and defense organization in which the author worked for many years since it was too complex and time-

The author is from the Faculty of Engineering, Bar-Ilan University, izack.cohen@biu.ac.il

The research was funded under ISF grant no. 226/21

Manuscript received April 19, 2021; revised August 16, 2021.

consuming for planners to manually analyze multiple previous projects.

We believe that the two main difficulties primarily associated with analogy-based planning approaches are: 1) Basing a new project plan on a specific previous project ignores other projects that may be a better starting point for the plan, and 2) a previous plan and project schedule embed hidden, temporal organizational constraints that would not necessarily be valid for the next project. These redundant constraints may restrict scheduling procedures from converging to optimal schedules. Two examples of types of redundant constraints are: 1) When an organization faces high demand, it might not have enough resources to execute multiple project activities simultaneously. This limitation forces sequential project planning rather than performing activities in parallel. When, however, scheduling a new project under different resource availability conditions, we aim to identify and alleviate these constraints to enable more concurrent activities, ultimately reducing project durations. 2) Specific project circumstances can introduce unique constraints on activity sequencing that, if generalized to other projects, might limit efficiency. For instance, in the case of a crack in an aircraft wing, the necessity to 'drain fuel from the wing' before conducting 'lower wing maintenance' might prolong the overall duration of these two activities, even though they could be carried out simultaneously in a shorter timeframe when the wing is in tact.

The conclusion to be drawn from these two examples is that, if not relaxed, such constraints unnecessarily limit the planning of a future project and may lead to longer than necessary durations and to sub-optimal resource allocations.

An additional difficulty derives from the fact that relying on a previous project plan 'hides' other possible project variations that may be more suitable for the new project.

The fourth industrial revolution [7], which spans our digital and physical worlds, is generating an abundance of event data that can be used for discovering, managing, and controlling processes. Accordingly, we contend that today, almost 70 years after its inception, project management is increasingly supported by information systems that facilitate project data collection and analyses. The available data can be used to solve some of the above-mentioned problems. Nevertheless, basic planning procedures such as defining activities, their precedence relations and schedules, still rely on manual work and do not fully utilize the existing data.

In this work we harness the power of data and process science to support project management – a combination that, according to previous studies that mapped the integration of data science techniques into project management as a knowledge source [8], is sorely lacking. More specifically, we apply a set of process mining [9] and machine learning techniques to support the decisions made by a planner regarding the next project's plan.

The suggested data-driven approach automatically reviews data from multiple previous projects to construct a project network that can be used for planning and scheduling a new project. For this, we model a project via Petri nets that include constructs such as AND and XOR splits and joins, and sequences of activities. Some of these constructs, such as XOR, which are not used in traditional project management models (e.g., activity on node (AON) and activity on arc (AOA) graphs), enable different project variations within a single network to emerge and be expressed. The proposed approach can save planning time and offer the planner flexibility in choosing a project plan from likely project realization options.

The main contributions of this work to the literature about project network planning and to decision makers are:

1) Theory and methodology: While there are mature techniques for resource-constrained project scheduling of a *given* project network, there is a gap in research about automated, data-driven approaches to prepare the project model (see [8]). This paper narrows this gap by suggesting an approach that supports project decision making and scheduling by harnessing the power of data science, machine learning, and process mining.

   By defining related process mining and project management concepts, tools from one domain can be used in the other. For example, data from past projects will be used to learn a project Petri net that serves to build a relaxed project model for a current project. This model can be analyzed easily using a linear mathematical program to find the critical path – that is, the shortest project duration without resource constraints, which is the basis for resource-constrained project scheduling.

   A Petri net, which captures multiple possible project variations in conjunction with their frequencies, can be used to distinguish between rare and frequent project variations and make the project network explainable. Differently from operational processes in which process mining is used to measure compliance or for process enhancement, here the focus is on process mining to assist decision making vis-à-vis the new project's plan and its resource-constrained schedule (see [10] and [12]).
   A high-level view of the suggested approach, its flow and associated tools and techniques, is presented in Figure 1.
2) Practice: We formalize the proposed methodology via an algorithm and demonstrate it, for the first time as far as we know, in the context of *project planning*.

   We illustrate possible benefits from the approach using a running example and two real-life project datasets that were collected and published in [9, 13]. We believe that the proposed approach can be applied to project planning using the suggested algorithm and available tools and software.

The study is structured as follows: The next section presents a running example that serves to motivate the approach and to illustrate its steps. Section III reviews the relevant literature. Then, Sections IV and V detail the steps of the proposed approach and formalize them, respectively. Section VI presents the experiments, and Section VII highlights theoretical and managerial implications. The last section concludes the paper and outlines future research directions.

## II. Motivating Example

Consider a typical event log that stores information about projects (e.g., apartment building projects, information systems development projects etc.). Each record typically includes a project-ID number, an activity/event name, a timestamp (e.g., start time) and associated duration. In projects, the data also include information about resources, costs, clients, performances etc. Table I presents an excerpt from an event log used for our running example after grouping by project-ID and ordering the events chronologically by their timestamps.

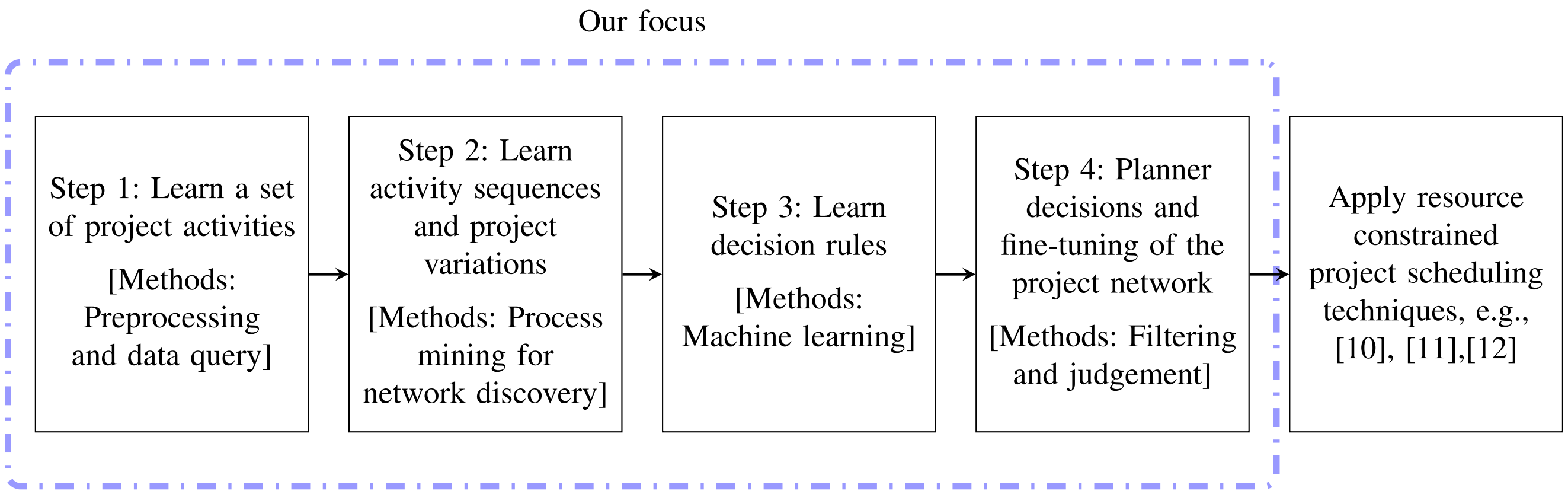

Figure 1. A diagram that provides a high-level view of the flow of the suggested approach and involved tools and techniques.

| Project-ID | Event-ID | Activity | Timestamp | Duration (h) | Client |
|---|---|---|---|---|---|
| 1 | $e_1$ | $a$ | 13-01-2022T12:00 | 2:00 | CO |
| 1 | $e_2$ | $b$ | 13-01-2022T14:55 | 4:00 | CO |
| 1 | $e_3$ | $c$ | 14-01-2022T08:39 | 3:30 | CO |
| 1 | $e_4$ | $e$ | 03-02-2022T11:47 | 5:00 | CO |
| 2 | $e_1$ | $a$ | 12-09-2020T11:07 | 2:15 | IZ |
| 2 | $e_2$ | $d$ | 20-09-2020T08:40 | 1:30 | IZ |
| 2 | $e_3$ | $e$ | 20-09-2020T11:32 | 4:30 | IZ |
| 3 | $e_1$ | $a$ | 10-12-2021T13:00 | 2:30 | TA |
| 3 | $e_2$ | $c$ | 28-12-2021T10:40 | 3:00 | TA |
| 3 | $e_3$ | $b$ | 10-01-2022T08:55 | 4:00 | TA |
| 3 | $e_4$ | $e$ | 13-02-2022T09:47 | 3:30 | TA |
| 4 | $e_1$ | $a$ | 10-11-2021T15:05 | 2:00 | IZ |
| 4 | $e_2$ | $d$ | 03-02-2022T11:40 | 1:30 | IZ |
| 4 | $e_3$ | $e$ | 05-02-2022T16:22 | 4:30 | IZ |
| ⋮ | ⋮ | ⋮ | ⋮ | ⋮ | ⋮ |

Table I
A SAMPLE EVENT LOG. EACH ORDERED PROJECT FORMS A TRACE.

Let us discuss the idea of learning from several projects instead of selecting one as our template. Consider, for example, a planner who selects Project 1 as a template for the next project. While our illustration in Figure 2(a) is intentionally simplistic, basing a new project on a template of a previous one is a reasonable practice since projects include dozens or hundreds of activities that make it almost impossible to manually analyze them, retaining what might be useful and discarding irrelevancies. The AON network of Project 1 depicts a sequential project where the minimal project duration (i.e., the critical path) can be found through the mathematical program in Equation 1. The formulation determines the activity start times, $S_i, \forall i \in \{a, b, c, d, e\}$ with the aim of minimizing project duration, which is $p_a + p_b + p_c + p_e$ ($p_i$ denotes the duration of activity $i$) with start times $\{S_a = 0, S_b = p_a, S_c = p_a + p_b, S_e = p_a + p_b + p_c\}$. The implicit assumption in Project 1, carried through to the next project plan, is that activities should be performed in succession because of, for example, physical constraints (e.g., a wall can be built only after the floor is finished) or resource limitations for this project (e.g., there are only two resource units available and each activity requires these two resource units).

$$\begin{aligned} \min_{S_i} \quad & S_e + p_e \\ \text{s.t.} \quad & S_b \geq S_a + p_a \\ & S_c \geq S_b + p_b \\ & S_e \geq S_c + p_c \\ & S_{start} \geq 0. \end{aligned} \tag{1}$$

Assume that we reveal, by analyzing several other projects, that activities $b$ and $c$ can actually be performed in parallel (e.g., there is no physical constraint between them). Consequently, the project network can be formulated as in Figure 2(b) and the minimal duration can be found using the following mathematical program (in Section IV we provide details regarding how to discover a relaxed project network):

$$\begin{aligned} \min_{S_i} \quad & S_e + p_e \\ \text{s.t.} \quad & S_b \geq S_a + p_a \\ & S_c \geq S_a + p_a \\ & S_e \geq S_b + p_b \\ & S_e \geq S_c + p_c \\ & S_{start} \geq 0. \end{aligned} \tag{2}$$

Equation 2 *relaxes* the precedence constraint between $b$ and $c$. Accordingly, the minimal duration of the relaxed formulation in Equation 2 is $p_a + \max\{p_b, p_c\} + p_e$; thus time is saved and, moreover, the model can also offer the flexibility to delay the start time of the activity associated with $\min\{p_b, p_c\}$ by $\max\{p_b, p_c\} - \min\{p_b, p_c\}$ without delaying the project completion (denoted as slack in project scheduling). The duration reduction from this relaxation can amount to $p_b + p_c - \max\{p_b, p_c\} = \min\{p_b, p_c\}$, assuming enough resources to schedule $b$ and $c$ in parallel.

Since the reduction in duration is monotonically non-decreasing with the number of relaxed constraints, the planner can enjoy increasing flexibility in generating project schedules. For example, consider sorting activities $1, \ldots, n$ in a decreasing order of duration, i.e., $p_1 = \max\{p_1, \ldots, p_n\}$. The potential duration reduction from modeling the activities in parallel compared to a sequential model is $\sum_{i=2}^{n} p_i$ (in Section VI, we present the potential duration reduction of a real-world project). To prepare the new project schedule, the planner has to take into account resource constraints (e.g., workers, cash flow, etc.) and use resource-constrained project scheduling techniques. We note that resource constraints are typically temporal, affected by the overall amount of organizational resources and the amount committed to other ventures during the new project planning horizon.

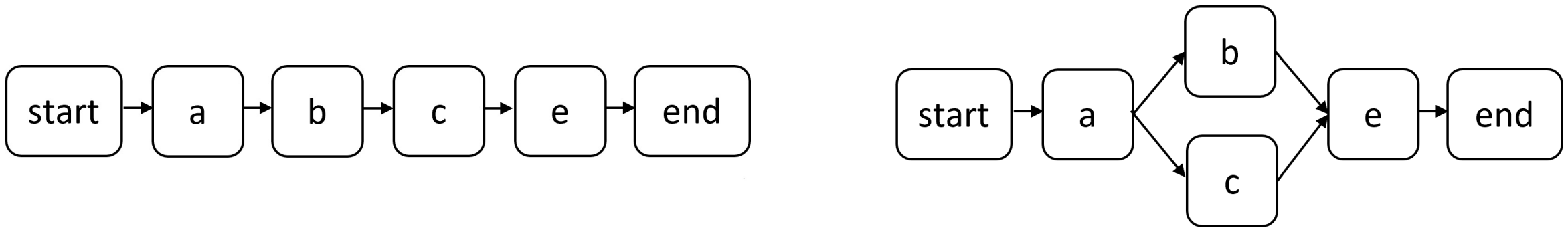

(a) (b)

Figure 2. Two AON networks that accommodate (a) Project 1 $[\langle a, b, c, e\rangle]$ and (b) Projects 1 and 3 $[\langle a, b, c, e\rangle, \langle a, c, b, e\rangle]$.

## III. Literature Review

The mainstream literature about planning a project network relies on time-consuming manual work that involves defining activities and sequencing them, typically done by experts (see [6]). The abundance of data recorded in information and project systems provides an opportunity to revolutionize project management as noted in [14]. Indeed, recent studies encourage using data-driven methods for project management. For example, Erfani et al. [15] use natural language processing techniques to identify risks in transportation projects.

This paper agrees with Bakici et al.'s [14] recommendation to enrich common project network planning practices by using data-driven methods. Accordingly, this is the focus of the literature review. Table II lists several data-driven methods that can be used in project management. Differently from other project management approaches, the suggested approach allows for the automatic discovery of a project network that takes into account previous project variations, relaxes project constraints to reduce the project duration, and identifies decision rules that can explain project variations.

| Ref | Resource Relaxation | Decision Rules | Model Discovery | Real Data |
|---|---|---|---|---|
| Erfani et al.,2023 | − | − | − | + |
| Joe et al.,2016 | − | − | + | − |
| Zebro and Timinger,2022 | − | − | + | − |
| Vavpotič et al.,2022 | − | − | − | + |
| Kouzari et al.,2023 | − | − | + | + |
| Urrea-Contreras et al.,2022 | − | − | + | + |
| This paper | + | + | + | + |

Table II
A HIGH-LEVEL VIEW OF RELATED DATA-DRIVEN APPROACHES FOR PROJECTS WITH RESPECT TO THE SUGGESTED APPROACH'S FEATURES.

Researchers agree that knowledge about how to integrate process mining techniques into project management is lacking (see the 2021 review by [8]). Some studies, such as [21], discuss information systems from which project data can be extracted without providing examples for the uses of such data. Despite the increasing importance of projects and the fundamentally different approach for their planning and management compared to operational processes, we found only a few articles that combine project management and process mining. We review them below.

In 2016, [16] suggested that data from previous recurring projects can be used to reveal insights about a project type using the heuristic miner [22], an idea that we follow. Nevertheless, that study and others that followed (e.g., [17]) did not highlight the aspects that we tackle such as relaxing resource constraints, revealing decision rules that can guide the selection of an appropriate project variation, and the added flexibility and potential improvement in project scheduling and resource allocation procedures. They likewise did not provide examples based on real project data.

One stream of research (see [19], [20]) investigated software development projects by applying process mining techniques using data from bug closure and issue tracking systems such as JIRA (e.g., [18]) and version control systems. These systems, however, cover only the problem solving and version control aspects of a project. Thus, they cannot be utilized to improve the project planning aspects on which we focus such as constraint relaxation in favor of resource allocation and duration optimization.

In summary, the current state-of-art in data-driven approaches for projects is mainly focused on model discovery. The contribution of this paper includes improving project planning and scheduling by discovering possible project variations, their associated decision rules, and constraint relaxation that may enable shorter project durations and efficient scheduling– features that are absent from the previous related research.

## IV. Modeling Approach

### A. Preliminaries

We denote a set of events and activities grouped by a project-ID as a *trace* – a chronologically-ordered sequence of events and activities $e_1, e_2, \ldots$ such that $t(e_j) \geq t(e_i), \forall j > i$, where $t(e_j)$ is the timestamp (typically, the start time) for activity $j$. Each trace represents a chronologically-ordered project realization (hereafter, we use the term project realization). Table I includes three types of project realizations for the four projects that compose the event log: $L = [\langle a, b, c, e\rangle, \langle a, c, b, e\rangle, \langle a, d, e\rangle^2]$, where $\mathcal{A}$ is a finite set of activities such that $\{a, b, c, d, e\} \in \mathcal{A}$. $\tau$ denotes a dummy activity that is not recorded in the log (e.g., when a project part is not recorded or a dummy activity is needed). Thus, the full set of activities is $\mathcal{A} \cup \tau$. The project realizations in Table I could be categorized into the three different sequences that appear in event log $L$.

In real-world settings, event logs include dozens of projects, each of which has dozens of activities, making manual network design hard. Accordingly, we propose an approach for automatically extracting a project model that compactly captures past realizations and includes information that can help a planner developing the next project's network.

### B. Modeling Languages

Project networks are typically modeled via precedence graphs such as AOA or AON. These graphs include activity sequences and AND splits and joins. Absent constructs such as exclusive choice (XOR) and inclusive choice (OR) mean that AOA and AON networks cannot be used for compactly capturing several project realizations within a single network. These can be captured, on the other hand, using Petri nets [23] and process tree representations [9].

A Petri net is a directed bipartite graph consisting of two types of nodes: *places* and *transitions*. Places are depicted as white circles, while transitions are represented by rectangles. The nodes are connected via directed arcs; connections between two nodes of the same type are not allowed. Places may contain zero or more tokens, which are depicted as black dots. The distribution of tokens over places describes the state of the Petri net. A place p is called an input place of a transition t if there exists a directed arc from p to t. Similarly, p is called an output place of t if there exists a directed arc from t to p.

Since our main target is to automatically learn and enrich a network from previous project realizations, we use process trees and Petri nets for which there are specialized learning algorithms.

Petri nets, by nature, do not support time and data, have non-deterministic transition firing, and transitions fire as soon as possible, which can limit our project planning approach. We deal with some of these limitations by using a *timed* Petri net model, which extends the standard Petri net, and by handling the data perspectives of projects via machine learning. Other limitations, which are less relevant for the domain of project network planning, are eclipsed by the advantages that Petri nets and associated process mining techniques provide for learning from previous projects. The mathematical foundations of Petri nets enable us to formally check network properties such as correctness and soundness that may be important in the context of project planning. For example, these checks enable the planner to verify that a project can be completed, that there are no dead parts within the network, etc.

We begin by defining a Petri net and a project tree.

**Definition 1** (Petri net; see [9] Definition 3.2). *A Petri net is a triplet $N = (P, T, F)$ where $P$ is a finite set of places, $T$ is a finite set of transitions (activities) such that $P \cap T = \emptyset$, and $F \subseteq (P \times T) \cup (T \times P)$ is a set of directed arcs, called the flow relations. A marked Petri net is a pair $(N, M)$, where $N = (P, T, F)$ is a Petri net and $M \in \mathbb{B}(P)$ is a multi-set of tokens over $P$ denoting the marking of the net. The set of all marked Petri nets is denoted $\mathcal{N}$.*

As an example, the Petri net equivalents of the AONs in Figure 2(a) and 2(b) are presented in Figure 3(a) and 3(b), respectively. The two Petri nets are marked – the black token indicates that the projects are in their start states.

In projects in which transitions typically correspond to activities, it is more appropriate to use *timed* Petri nets [24, 25]. A timed Petri net, in our case, extends the standard Petri net by associating places or transitions with time to reflect activity durations. Thus, tokens have an age, representing the time since their creation. When a timed transition fires, it increases the age of each token by a specific real number. Essentially, the definition of a timed Petri net (which is excluded for compactness) is based on a marked Petri net as defined in Definition IV-B with a firing time function that assigns a positive rational number to each transition. As in a standard Petri net, a transition must be enabled to start and then takes a positive amount of time to be performed. This reflects project dynamics – to start execution, an activity's precedence relations have to be satisfied, and upon its start, an activity is processed according to its duration.

Petri nets can be transformed into project trees and vice versa and each model has its related network discovery algorithms. Thus we define a project tree as follows.

**Definition 2** (Project (process) tree; see [9] Definition 3.13). *Let $A \subseteq \mathcal{A}$ be a finite set of activities with $\tau \notin A$.*
*$\oplus = \{\rightarrow, \times, \wedge, \circlearrowleft\}$ is the set of project tree operators.*

- *If $a \in A \cup \{\tau\}$, then $Q = a$ is a project tree,*
- *if $n \geq 1$, $Q_1, Q_2, \ldots, Q_n$ are project trees, and $\oplus = \{\rightarrow, \times, \wedge\}$, then $Q = \oplus(Q_1, Q_2, \ldots, Q_n)$ is a project tree, and*
- *if $n \geq 2$ and $Q_1, Q_2, \ldots, Q_n$ are project trees, then $Q = \circlearrowleft (Q_1, Q_2, \ldots, Q_n)$ is a project tree.*

A project tree includes four types of operators: $\oplus = \{\rightarrow, \times, \wedge, \circlearrowleft\}$, where $\rightarrow$ marks sequential composition, $\times$ denotes exclusive choice, $\wedge$ is a parallel composition, and $\circlearrowleft$ is a redo loop for repetitions of project parts.

Using definitions IV-B, and IV-B, Figure 3 presents three logs, with their related AONs, and Petri nets. Figure 4 presents a project tree for the running example (Section IV-C explains how to discover the process tree from the log). For our simple running example, one can see that the Petri net in Figure 3(c) and the project tree in Figure 4 accommodate all three project variations (realization patterns) while an AON model is more limited and cannot capture the set of all three types of projects from Table I. The former models are associated with algorithms that facilitate automatic learning from an event log, which makes them especially suitable for our needs. In the next section we present one such learning approach.

### C. Learning a Project Model

We aim to learn a project model from an event log of past projects. Process mining offers several model learning algorithms such as inductive mining (IM), fuzzy miner, heuristic miner, ILP-based algorithms, genetic miner and more (see [9]).

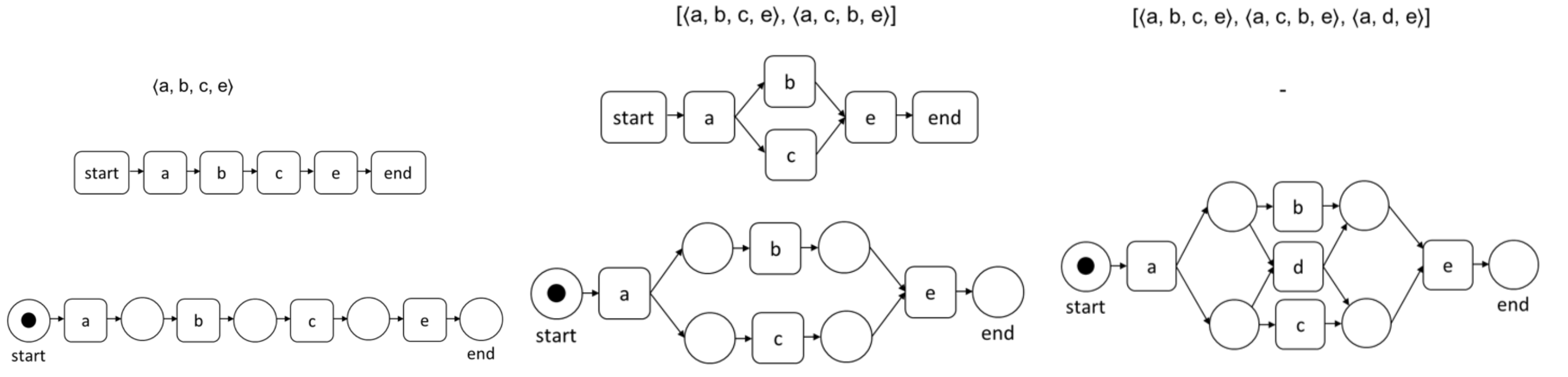

(a) (b) (c)

Figure 3. AON, Petri net and project tree models (from top down, respectively) for three project realization logs: (a) $L = [\langle a,b,c,e\rangle]$, (b) $L = [\langle a,b,c,e\rangle, \langle a,c,b,e\rangle]$, and (c) $L = [\langle a,b,c,e\rangle, \langle a,c,b,e\rangle, \langle a,d,e\rangle]$. Note that an AON model cannot model the log.

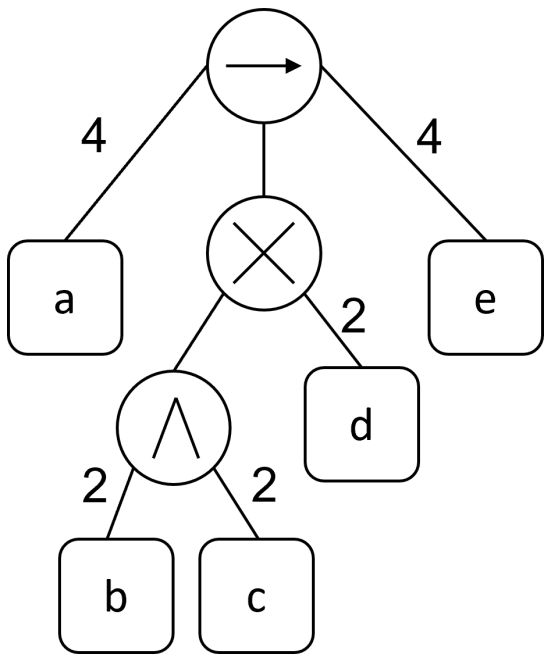

Figure 4. The project tree for the running example.

In this paper, we use the IM algorithm since it can handle large logs while ensuring formal properties such as correctness and the ability to rediscover the source model. Some of the listed weaknesses of IM such as its generalization ability, reliance on directly-follows graphs (DFGs), and frequency-based relations are actually a benefit in the context of project planning, as we will elaborate on later in this paper. We acknowledge that alternative discovery algorithms are available but exploring them is outside the scope of this paper (for a review of contemporary discovery methods refer to [26]). Next, we present the main principles of the IM algorithm [27, 28] and illustrate them using the running example.

IM discovers a tree that can be transformed easily into a Petri net model and vice versa. Petri nets form mathematically sound, rich network representations and include constructs such as AND, exclusive choice (XOR), loops, and execution semantics that enable model verification. IM is used to learn a project model from historical realizations, which makes it qualify as a major component within the proposed automatic data-driven project planning approach.

IM recursively splits a log $L$ into smaller and smaller sub-logs by applying four types of cuts that represent the operators $\{\rightarrow, \times, \wedge, \circlearrowleft\}$: $\rightarrow$ sequence, $\times$ exclusive choice, $\wedge$ parallel composition, and $\circlearrowleft$ redo loop. Each sub-log, which includes a set of sub-traces, is split again until each sub-trace includes a single activity.

The first step identifies links between activity couples that directly follow each other in the different traces in favor of constructing the DFG for the project log. We note that observing a project in which one activity directly follows another activity is a necessary but not sufficient criterion to establish a predecessor–successor relationship between them since the two activities may be concurrent; for example, in the running example of Table I in Project 1, $b \rightarrow c$ but this observation does not constitute a predecessor–successor relationship since in Project 3, $b \rightarrow c$.

First, let us define a DFG.

**Definition 3** (Directly-follows graph)**.** *A DFG is a pair* $G = (A, F)$ *where* $A \subseteq \mathcal{A}$ *is a finite set of activities,* $\blacktriangleright, \blacksquare \notin A$ *are dummy start and end nodes, respectively, and* $F \in (A \times A) \cup (\blacktriangleright \times A) \cup (A \times \blacksquare) \cup (\blacktriangleright \times \blacksquare))$ *is a multi-set of arcs.*

Figure 5 presents the DFG for the event log in Table I – $L = [\langle a,b,c,e\rangle, \langle a,c,b,e\rangle, \langle a,d,e\rangle^2]$.

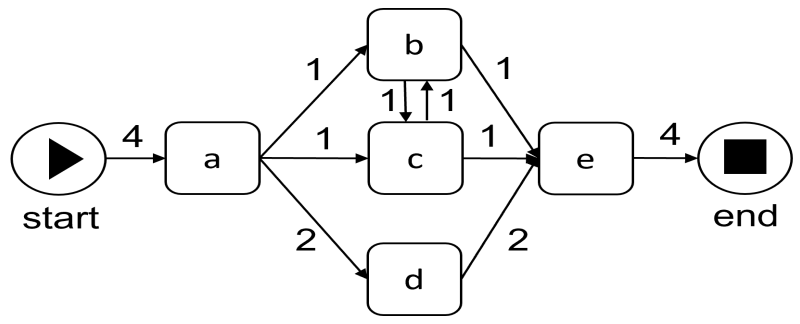

Figure 5. DFG of $L = [\langle a,b,c,e\rangle, \langle a,c,b,e\rangle, \langle a,d,e\rangle^2]$. Numbers denote frequencies.

Let us mathematically define the cuts applied on a DFG that was constructed based on event log $L$ (e.g., [27]). We illustrate some of the cuts using our running example.

**Definition 4** (Cuts of DFG)**.** *Given a DFG for event log* $L$, $G(L) = (A, F)$*, an* $n$*-degree cut* ($n \geq 1$) *partitions* $L$ *into* $n$ *disjoint sets of activities* $A_1, A_2, \ldots, A_n$ *such that* $A_L = \cup_{i \in \{1,\ldots,n\}} A_i$ *and* $A_i \cap A_j = \emptyset \; \forall i \neq j$.

*There are four types of cuts, each of which corresponds to one project tree operator* $\oplus = \{\rightarrow, \times, \wedge, \circlearrowleft\}$*, where* $\rightarrow$ *marks sequence composition,* $\times$ *denotes exclusive choice,* $\wedge$ *is a parallel composition, and* $\circlearrowleft$ *is a redo loop for repetitions of project parts. The conditions for defining each cut of* $G(L)$ *are:*

- *A sequence cut, denoted by* $(\rightarrow, A_1, A_2, \ldots, A_n)$*, satisfies* $\forall i, j \in \{1, \ldots, n\} \forall a \in A_i \forall b \in A_j \; i < j \Rightarrow a \mapsto^+$

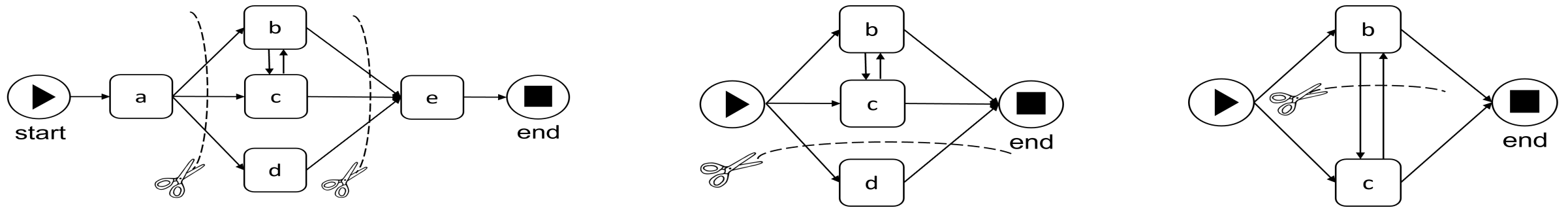

(a) (b) (c)

Figure 6. Three cuts made by the IM algorithm performed on respective DFGs of the running example: (a) A sequence cut $\rightarrow$ on log $[\langle a,b,c,e\rangle, \langle a,c,b,e\rangle, \langle a,d,e\rangle^2]$, (b) an exclusive-choice cut $\times$ on the sub-log $[\langle b,c\rangle, \langle c,b\rangle, \langle d\rangle^2]$, and (c) an AND cut $\wedge$ on sub-log $[\langle b,c\rangle, \langle c,b\rangle]$.

*$b \wedge b \not\mapsto^+ a$, where $a \mapsto^+ b$ denotes that the DFG includes a non-empty path from $a$ to $b$.*

- *An exclusive-choice cut, denoted by $(\times, A_1, A_2, \ldots, A_n)$, satisfies $\forall i,j \in \{1,\ldots,n\} \forall a \in A_i \forall b \in A_j \; i \neq j \Rightarrow a \not\to b$.*
- *A parallel cut, denoted by $(\wedge, A_1, A_2, \ldots, A_n)$, satisfies:*
  - *$\forall i \in \{1,\ldots,n\} \; A_i \cap A^{start} \neq \emptyset \wedge A_i \cap A^{end} \neq \emptyset$, where $A^{start}, A^{end}$ are the sets of start and end activities in L, respectively, and*
  - *$\forall i,j \in \{1,\ldots,n\} \forall a \in A_i \forall b \in A_j \; i \neq j \Rightarrow a \to b$.*
- *A redo loop cut, denoted by $(\circlearrowleft, A_1, A_2, \ldots, A_n)$, satisfies:*
  - *$n \geq 2$,*
  - *$A^{start} \cup A^{end} \subseteq A_1$,*
  - *$\{a \in A_1 | \exists i \in \{2,\ldots,n\} \exists b \in A_i \; a \to b\} \subseteq A^{end}$,*
  - *$\{a \in A_1 | \exists i \in \{2,\ldots,n\} \exists b \in A_i \; b \to a\} \subseteq A^{start}$,*
  - *$\forall i,j \in \{2,\ldots,n\} \forall a \in A_i \forall b \in A_j \; i \neq j \Rightarrow a \not\to b$,*
  - *$\forall i \in \{2,...,n\} \forall b \in A_i \exists a \in A^{end} \; a \to b \Rightarrow \forall a' \in A^{end} \; a' \to b$, and,*
  - *$\forall i \in \{2,...,n\} \forall b \in A_i \exists a \in A^{start} \; b \to a \Rightarrow \forall a' \in A^{start} \; b \to a'$.*

  *A cut $(\oplus, A_1, A_2, \ldots, A_n)$ of $G(L)$ is maximal if there is no other cut $(\oplus, A_1, A_2, \ldots, A_m)$ with $m > n$.*

For the running example, the first cut, illustrated in Figure 6(a), is the sequence cut ($\rightarrow$) that splits the log into three sub-logs $[\langle a\rangle^4]$, $[\langle b,c\rangle, \langle c,b\rangle, \langle d\rangle^2]$, and $[\langle e\rangle^4]$. Two of the sub-logs are singletons and cannot be split further.

The next IM cut for sub-log $[\langle b,c\rangle, \langle c,b\rangle, \langle d\rangle^2]$ is the exclusive-choice cut ($\times$), as can be seen in Figure 6(b). The resulting sub-logs are $[\langle a\rangle^4]$, $[\langle b,c\rangle, \langle c,b\rangle]$, $[\langle d\rangle^2]$ and $[\langle e\rangle^4]$. Again, two of the sub-logs are singletons and cannot be split further.

The final cut splits the sub-log $[\langle b,c\rangle, \langle c,b\rangle]$ using the AND ($\wedge$) cut as presented in Figure 6(c). At this point, all sub-logs are singletons. The resulting process tree is presented in Figure 4. Note we can use the frequencies of the sub-logs that are marked as superscripts (e.g., $[\langle e\rangle^4]$ indicates that $e$ happened four times) to enrich the project tree with additional information. In Section IV-D we show how a planner can use these frequencies to filter out rare project variations. The enriched project tree in Figure 4 can easily be represented as an enriched Petri net.

### D. Deciding on the Project Model

The output of Section IV-C is a project tree or a Petri net that accommodates a variety of possible project realizations. In our running example, the project tree in Figure 4 represents three possible realizations: $\langle a,b,c,e\rangle$, $\langle a,c,b,e\rangle$, which represent the same type of project in which $a$ precedes $b$ and $c$ that can be done in parallel, and $\langle a,d,e\rangle$. Differently than operational process that are not explicitly scheduled, when planning a project a specific variation (a model path) must be selected as the project plan, which is used for allocating resources, deciding on a schedule linked to payment milestones etc. To help the planner in choosing a project variation, we augment the model with decision rules at exclusive-choice splits and joins, and by filtering out rare project variations based on their frequency.

*1) Filtering by Frequencies:* Simplifying a project model by filtering can be done in several ways such as not considering less frequent project activities, less frequent project variations (activity sequences) or arcs in the DFG. Distinguishing between less and more probable network paths has been studied in the context of project management (see [5],[29]) and in the context of process mining ([28],[30],[31],[32]). The approach we take is to construct a model based on the complete event log and then filter out paths with a 'slower' flow according to a specified threshold. In other words, the planner eliminates project variations that are considered rare. We illustrate the idea using the running example. Assume that the complete event log in Table I includes 100 projects that can be represented as $L = [\langle a,b,c,e\rangle^{45}, \langle a,c,b,e\rangle^{53}, \langle a,d,e\rangle^2]$.

As noted in Section IV-C, it is easy to uncover a project tree annotated with frequencies and represent it as a Petri net. For illustration, in Figure 7(a)), we present the frequency-enriched Petri net that was learned from the running example. Assuming that a planner wants to eliminate rare project variations by using a filter of $5\%$ of the cases, we get the reduced model presented in Figure 7(b)), which does not contain activity $d$. For realistic models, the number of variations can be high; thus, filtering can enable the planner to focus on project variations deemed more important.

*2) Explaining the Model:* Project datasets include much more than the basic details needed for learning a network. Typically, there are project-level features such as the client's name, budget details, and manager's name, and activity-level features such as durations, start and completion dates, cash inflows and outflows, the types and amounts of the required resources, and more. A project can be presented as a feature vector and machine-learning techniques such as regression, decision trees and deep learning networks can be used to explain a selected label and to generate predictions of values of interest.

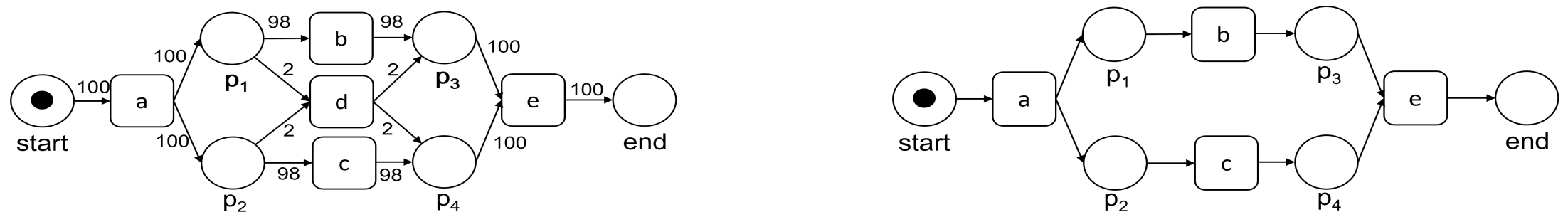

Figure 7. Petri nets for $L = [\langle a, b, c, e\rangle^{45}, \langle a, c, b, e\rangle^{53}, \langle a, d, e\rangle^{2}]$ (a) A model with frequencies, and (b) a reduced model with a 5% filter. Places are labeled start, $p_1, p_2, p_3, p_4$, end.

| **Project-ID** | **Feature-1** | **Feature-2** | $\cdots$ | $\{b, c\}$ |
|---|---|---|---|---|
| 1 | CO | \$ 50,000 | $\cdots$ | TRUE |
| 2 | IZ | \$ 10,000 | $\cdots$ | FALSE |
| 3 | TA | \$ 85,000 | $\cdots$ | TRUE |
| 4 | IZ | \$ 10,000 | $\cdots$ | FALSE |
| ⋮ | ⋮ | ⋮ | ⋮ | ⋮ |

Table III
EXAMPLE DATA FOR PREDICTING WHETHER THE CLASS LABEL IS $\{b, c\}$.

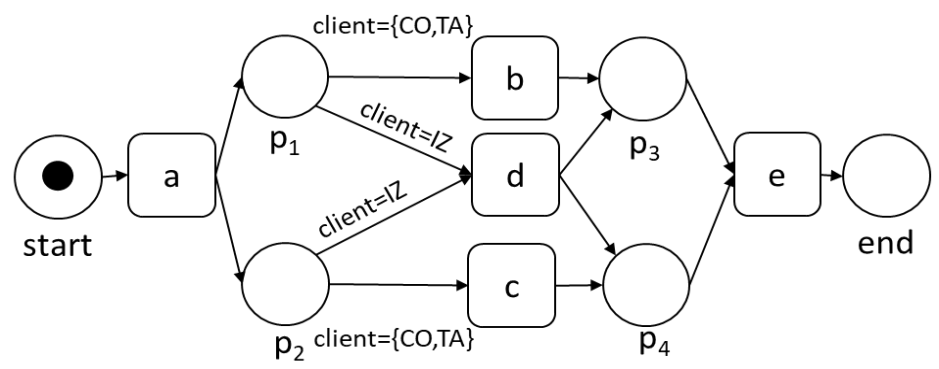

Figure 8. The Petri net of the running example with decision rule information. Places are labeled start, $p_1, p_2, p_3, p_4$, end.

Explainable models is an active research area in machine learning (see the paper by Singer and Cohen [33] on explainable decision trees). This idea was denoted in process mining as decision mining [34] or revealing guards [35]. Guards are decision rules that determine if, in a given process state, the data variables will allow a transition to become enabled. Contrary to the standard way in which decision rules are used in processes to determine the next process branch from several alternatives (e.g., a XOR split), in the present paper they function as an aid for planners for selecting a single alternative branch at XOR splits (and joins) to include in the network configuration of a new project.

We illustrate the idea using the running example. Table I includes supervised data that can be used for learning the Petri net in Figure 7(a) and for training and validation of a machine-learning model that learns decision rules. For the running example, places $p_1$ and $p_2$, each of which has two exclusive output branches, constitute a decision point. Learning the decision point is equivalent to identifying the conditions under which either the set of activities $\{b, c\}$ or activity $d$ would be realized. The key idea is to re-arrange the data such that the predicted class label would be either the set $\{b, c\}$ or $\{d\}$ after activity $a$, and the independent variables are selected data features. We illustrate such a data arrangement in Table III. For the running example, it is easy to see that if client="IZ", then $d$ and otherwise $\{b, c\}$. Most cases are more involved but nonetheless it is simple to apply standard classification or regression machine-learning models to identify decision rules. Tagging a model as shown in Figure 8 can help the planner decide on the new project's configuration.

## V. FORMALIZATION OF THE APPROACH

Algorithm 1 formalizes the suggested planning framework. The algorithm's input is a dataset $D$ that contains execution data about a class of organizational projects. Examples of project classes include Boeing 767 aircraft passenger-to-cargo conversion projects or apartment building project, among various others. There are several hyperparameters that can be set to a value or iteratively altered by the planner. The frequency threshold parameter, $\gamma$, controls how much noise is filtered. Choosing a value of $0.2$, for example, will result in keeping only project paths in which more than $20\%$ of the traffic flows. Higher $\gamma$ values amount to keeping only the most frequent project variations. Another parameter that the planner can choose is whether to extract decision rules – done by setting $d$ to 1. The dataset, $D$, is initialized to an event log structure – that is, to a multi-set of chronologically-ordered project executions. Then, a learning algorithm is applied to $L$ to learn a project tree $Q$ (Line 1) that can be represented as a Petri net model $N$ (Line 2), which is the starting point for further analyses. We note that we learn project models using IM. It is, however, worth mentioning that planners have the flexibility to choose alternative learning models.

A model refinement procedure is defined in Lines 3–14 for a planner who wants to refine $N$ and see its highways ($\gamma > 0$). The model's flow relations are scanned (Line 4) and each flow is annotated with its corresponding frequency $f(e)$ (Line 5) – how to extract the corresponding frequencies easily is explained in the last paragraph of Section IV-C. Essentially, the threshold is translated into traffic conditions (Line 6) and flow relations that do not meet the threshold are filtered out (Line 7).

Removal of flow relations may create unconnected activities that need to be removed. We denote unconnected activities as those that have empty sets of input and output places $\bullet$t and t$\bullet$, respectively, and remove them in Line 10. Likewise, we remove unconnected places (Line 11). Finally, the refined project model is returned (Line 13).

The algorithm is designed to use the filtered model (for $\gamma > 0$) for decision rule learning, when $d = 1$ (Line 15), but it can also use the unfiltered model. Decision rules are stored in a set, $Rules$, of tuples $(r, d_r)$, where $r$ is a decision point and $d_r$ is its respective decision rule. First, $Rules$ is set to an empty set (Line 16). Next, decision points, which are exclusive choice points (places) with two or more output activities, are mapped into set $R$ (Line 17). $R$ may include many decision points; thus, a planner may prefer to learn only a sub-set, $R^{'}$, of decision points that they deem more important (Line 18). For each selected decision point (Line 19), $D$ is arranged to facilitate the use of a machine-learning algorithm with selected features (see Table III and Line 20). In Lines 21–22, a rule is learned and added to the set of rules. Lastly, the project network and the set of rules are returned.

**Algorithm 1** Data-driven project planning

**input** : project dataset $D$, the number of recorded projects $n$, hyperparameters: a frequency threshold $\gamma \in [0, 1)$ (0 means no filtering), decision rule learning $d \in \{0, 1\}$ (0 for not considering decision rules)

**output** : a filtered project Petri net $N$, a set of tuples $(r, d_r)$, where $r \in R^{'}$ is a set of selected decision points, $d_r$ is the decision rule for $r$ and their rules $Rules$

**initialization:** represents dataset $D$ as an event log $L$.

1 learn a project tree $Q$ that corresponds to $L$
`// apply the inductive miner (see Section IV-C)`
2 represent the project tree as a Petri net $N = (P, T, F)$
3 **if** $\gamma > 0$ **then**
`// filtering, see Section IV-D1`
4 **for *each flow relation*** $e \in F$ **do**
5 annotate $e \in F$ with its frequency $f(e) \in \mathbb{N}$
6 **if** $f(e) < \lceil n \cdot \gamma \rceil$ **then**
7 $F \leftarrow F \setminus e$ `// filter out`
8 **end**
9 **end**
10 $T \leftarrow \{\text{t} | \bullet \text{t} \wedge \text{t}\bullet \neq \emptyset\}$ `// remove unconnected activities`
11 $P \leftarrow \{\text{p} | \bullet \text{p} \wedge \text{p}\bullet \neq \emptyset\}$ `// remove unconnected places`
13 return Petri net $N = (P, T, F)$
14 **end**
15 **if** $d = 1$ **then**
`// learning decision rules, see Section IV-D2`
16 $Rules = \emptyset$ `// set of tuples of decision points and decision rules` $(r, d_r)$
17 $R = \{\text{p} \in P | |\text{p}\bullet| > 1\}$ `// places with two or more outgoing flow relations`
18 select a subset of relevant decision points $R^{'} \subseteq R$
19 **for *each*** *decision point* $r \in R^{'}$ **do**
20 arrange $D$ as a vector with selected features
21 learn $r$ and produce $d_r$
22 $Rules \leftarrow (r, d_r)$
23 **end**
24 return Petri net $N = (P, T, F)$ and $Rules$
25 **end**

The planner now has an enriched model that captures activities, which can be performed in parallel, relevant project variations, and decision rules. This model is the starting point for performing resource-constrained project scheduling.

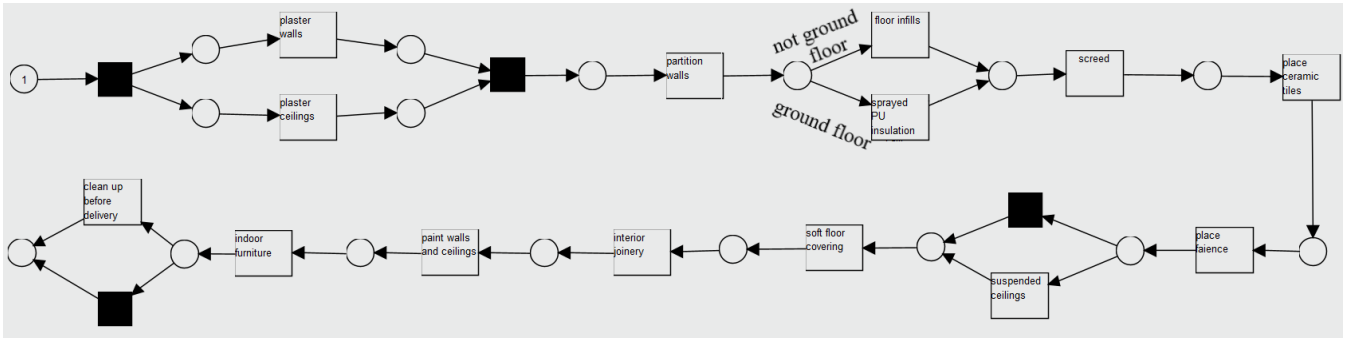

Figure 9. A Petri net of apartment finishing projects with 16 possible variations. Black transitions indicate $\tau$ activity – that is, no activity or a dummy activity.

## VI. Experiments

Datasets of real-world recurring projects record sensitive commercial and procedural information, which is why such datasets are typically not publicly available. We demonstrate the suggested approach by using the only publicly available real-world database, to the best of our knowledge, of recurring projects that has the required information (see Batselier and Vanhoucke [13] and Vanhoucke et al. [36]). This database is an ongoing initiative led by Prof. Mario Vanhoucke et al., continuously expanding with the addition of new projects.

### A. Data and Preprocessing

We used the finishing projects dataset to illustrate model construction, constraint relaxation, and making the model explainable. Then, we used data about residential homes to demonstrate a more complex project type and the magnitude of possible flexibility gains, in terms of possible duration reductions. These two datasets include a collection of apartments being finished and residential home building projects that were performed between 2015–2017. Each dataset details many project attributes such as activity names, start dates and durations, costs and resources. Preprocessing included standardizing activity labels such that a similar activity will have the same label across projects and arranging the dataset into an event log format. Once this was done, we applied IM for learning a project network and a classification decision tree for revealing decision rules. For our experiments, we used an Altair software tool – the RapidMiner, and Python with the Pm4py package.

### B. Discussion of Analyses and Results

The apartment finishing project model, revealed by applying the IM algorithm, is structured in the sense that the projects are relatively serial with a small amount of concurrency. Overall, the learned Petri net accommodates 16 possible project variations, as can be seen in Figure 9.

Next, we applied a classification decision tree to make the model explainable by learning exclusive choices that can guide the planner in selecting a specific project variation for the next planned project. For example, the exclusive choice between the 'floor infills' and 'sprayed PU insulation' activities is decided by whether the apartment under work is on the ground floor or not. The Petri net in Figure 9 presents the learned decision rule on the arcs that lead to the activities. Obviously, such decision rules can guide the selection of activities and their sequencing – in this case, including a floor insulation activity in ground floor apartments.

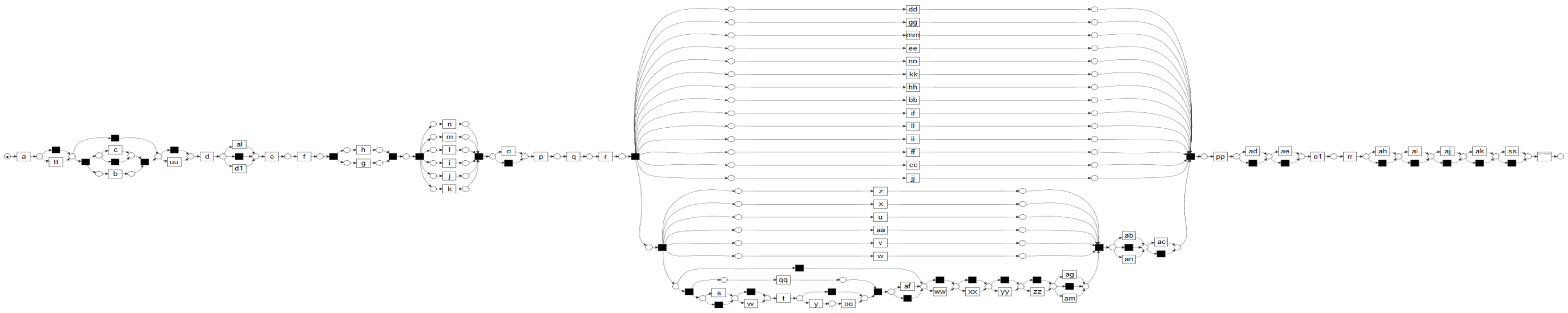

Figure 10. The learned Petri net of residential house construction projects. Black transitions indicate a $\tau$ activity – that is, no activity or a dummy activity.

Next, we developed a model of residential house construction projects. The corresponding Petri net is depicted in Figure 10. The model relaxes multiple constraints, which cannot be uncovered by inspecting an individual project, to show the activities that can now be done in parallel. For example, the baseline duration of project $2016 - 11$, which is 241 days, could be shortened to 178 days using the developed project model when considering the same planned activity durations. This consideration translates into a significant amount of activity slack and thus to resource allocation flexibility and potential shortening of a project's duration. While the learned project relaxed many constraints that were implemented in individual project plans due to temporal constraints and even though resource constraints should be considered when preparing the project schedule, the planner has much more flexibility owing to the additional 63 days that were stripped from the past project's resource constraints.

## VII. Theoretical and Managerial Implications

This paper models non-unique projects as processes with similar activity sequence patterns, and shared decision rules and resources. Such a view enables marrying project planning, process mining, and machine learning — a new paradigm in project management. The suggested data-driven project planning framework is a first step toward automated project planning, offering a project planner enhanced capabilities by automatically learning from past projects, revealing decision rules, and relaxing resource constraints.

Managers can use this approach to improve the efficiency and effectiveness of project planning and scheduling, which can lead to better project outcomes and higher customer satisfaction. The proposed approach can also help managers identify potential risks and opportunities in the project plan and schedule, which can lead to better decision making and resource allocation.

The approach offers other benefits as well. The different variations embedded in an enriched project network can assist planners trying to determine or predict the next project variation. Moreover, discovering project shortening opportunities can increase the competitiveness and the throughput rate of projects.

Equally important, the approach can easily be implemented using organizational information about previous projects and standard process mining and machine learning tools. For example, organizations can use the popular Python programming language that includes process mining and machine learning packages or other tools that do not require programming knowledge (e.g., RapidMiner by Altair). Some of the highlighted ideas can be used in project planning practices even before implementing the approach; for example, planners can rely on several similar projects to plan the next project rather than rely on a single previous project. Implementing the approach for the first time can reveal significant improvement opportunities. Nevertheless, as more data are accumulated, it is worthwhile to rerun the approach to discover additional improvement opportunities hidden in the new data.

## VIII. Conclusions

We propose a data-driven project planning approach that uses historical projects' records in conjunction with process mining and data science techniques. The approach combines learning a project network from previous similar projects and enriching the network with information about probable paths and decision rules.

The approach, which examines and learns from multiple similar projects, enables the relaxing of constraints imposed on individual projects due to temporal resource constraints or specific project circumstances that dictated activity sequences in past projects. It also uncovers a variety of project configurations from which one should be selected as the plan for a new project. Relieving constraints necessarily shortens the critical path (by 26% for a real project), thus enabling the planner to shorten the project when applying resource-constrained scheduling. This is the first time, to the best of our knowledge, that a real-world project dataset is used to demonstrate data-driven project network planning. The suggested approach integrates project planning and data science techniques. As a last stage, common resource-constrained project scheduling approaches can be applied to the relaxed project network to decide on the project schedule.

A limitation of the suggested approach is that it can be implemented only in organizations in which project data are available. The absence of publicly available datasets on recurring projects restricts this research from estimating the extent of flexibility gains across various projects from differ-ent domains. Nevertheless, the constraint relaxation approach ensures the realization of such gains.

As future research directions, we propose to: 1) Develop additional real-world recurring project datasets from different domains such as healthcare, engineering projects, pharmaceutical drug development etc., 2) develop data-driven artificial intelligence-based scheduling mechanisms, and 3) extend the approach into project control to complement common project control mechanisms such as the *earned value* model.

**Izack Cohen** is a researcher (associate professor) in the Faculty of Engineering at Bar-Ilan university. His academic education is from the Technion – Israel Institute of Technology. Izack's current research spans process modeling, optimization, and data-driven decision-making.